\documentclass[final]{cvpr}

\usepackage{times}
\usepackage{epsfig}
\usepackage{graphicx}
\usepackage{amsmath}
\usepackage{amssymb}

\usepackage[pagebackref=true,breaklinks=true,colorlinks,bookmarks=false]{hyperref}

\def\cvprPaperID{50} 
\def\confYear{CVPR 2021}

\begin{document}

\title{MarkerPose: Robust Real-time Planar Target Tracking for Accurate Stereo Pose Estimation}

\author{Jhacson Meza \qquad Lenny A. Romero \qquad Andres G. Marrugo\\
Universidad Tecnológica de Bolívar, Cartagena, Colombia\\
{\tt\small \{jmeza, lromero, agmarrugo\}@utb.edu.co}
}

\maketitle

\begin{abstract}
   Despite the attention marker-less pose estimation has attracted in recent years, marker-based approaches still provide unbeatable accuracy under controlled environmental conditions. Thus, they are used in many fields such as robotics or biomedical applications but are primarily implemented through classical approaches, which require lots of heuristics and parameter tuning for reliable performance under different environments. In this work, we propose MarkerPose, a robust, real-time pose estimation system based on a planar target of three circles and a stereo vision system. MarkerPose is meant for high-accuracy pose estimation applications. Our method consists of two deep neural networks for marker point detection. A SuperPoint-like network for pixel-level accuracy keypoint localization and classification, and we introduce EllipSegNet, a lightweight ellipse segmentation network for sub-pixel-level accuracy keypoint detection. The marker's pose is estimated through stereo triangulation. The target point detection is robust to low lighting and motion blur conditions. We compared MarkerPose with a detection method based on classical computer vision techniques using a robotic arm for validation. The results show our method provides better accuracy than the classical technique. Finally, we demonstrate the suitability of MarkerPose in a 3D freehand ultrasound system, which is an application where highly accurate pose estimation is required. Code is available in Python and C++ at \url{https://github.com/jhacsonmeza/MarkerPose}.
\end{abstract}

\section{Introduction}
Planar markers or fiducial markers are useful in robotics, computer vision, medical imaging, and related fields for 6-degree-of-freedom (DoF) pose estimation, mapping, localization, and other pose-related tasks. Tracking an object's pose by attaching a marker to it and using a vision system is attractive because it can be highly accurate while still being a low-cost alternative when compared to other strategies~\cite{brown2018design}. For instance, marker-less techniques are not as robust or fast or not used for precise pose estimation~\cite{nath2019using}. Moreover, marker-based methods are useful in different environments. Robustly detecting these targets under different lighting or motion conditions is essential for ensuring reliable system operation and accurate tracking for spatial registration of acquired data~\cite{kam2018improvement, kim2020marker}.

There are several applications where highly accurate pose estimation is crucial. For example, for robot path planning and navigation~\cite{lee2018path}, Unmanned Aerial Vehicle (UAV) refueling~\cite{sun2019bionic}, or for surgical instrument guidance~\cite{basafa2017visual}. These applications require a robust marker detection method under different environments and lighting conditions to register accurate poses. Furthermore, motion blur robustness is highly desired for real-time operation.

In this paper, we propose MarkerPose: a robust, real-time pose estimation method using a planar target with concentric circles. Our method is inspired by Deep ChArUco~\cite{hu2019deep} and consists of two main stages. In the first stage, with a SuperPoint-like network, the circles' centers are estimated at a pixel level. In the second stage, the points are estimated at sub-pixel accuracy with a proposed lightweight encoder-decoder network called EllipSegNet. The network segments the marker ellipses for sub-pixel center estimation. The pair of stereo images are processed with the proposed method, and through triangulation, the marker's pose is determined. The experiments show the highly accurate results of MarkerPose compared with a classical sub-pixel center detection method~\cite{Romero:2020gp}. Also, we show the potential and suitability of our proposed method for pose estimation under low lighting conditions and high motion blur.

Our main contributions are:
\begin{enumerate}
    \item Real-time, highly accurate pose estimation based on triangulation using a marker of circles and stereo images.
    \item EllipSegNet: a lightweight network for ellipse segmentation meant for real-time applications.
    \item Sub-pixel marker points detection at full resolution through ellipse fitting, avoiding detection errors by detecting in a lower resolution and rescaling to the original.
\end{enumerate}

\section{Related Work}
Deep Convolutional Neural Networks (CNN) have been extensively used for object detection with, for example, Faster R-CNN~\cite{ren2016faster} or YOLO~\cite{redmon2016you}. Additionally, there are different examples for object segmentation like Mask-RCNN~\cite{he2017mask} or DeepLab~\cite{chen2017deeplab}. Detection in challenging conditions has also been addressed. For example, Rozumnyi~\etal~\cite{rozumnyi2020fmodetect} tackle fast-moving object detection in challenging high-movement conditions. However, these methods are not readily applicable for 6DoF object pose estimation. Alternatively, other works have addressed object detection for marker-less pose estimation. For example, Nakako~\cite{nakano2020stereo} proposed a stereo vision for pose estimation using a robot manipulator. Xiang \etal~\cite{xiang2017posecnn} proposed to jointly segment and estimate different objects' poses. Wang \etal~\cite{wang2019normalized} addressed pose estimation with depth information using an RGB-D camera. Nevertheless, these approaches are limited to moderately accurate pose estimation applications.

Pose estimation can be approached via keypoint detection. There are multiple examples of keypoint detection with CNNs for marker-less pose estimation. Human body pose estimation~\cite{andriluka2018posetrack}, hand pose estimation~\cite{ge20193d} or head pose estimation~\cite{gupta2019nose} are popular examples in the computer vision community where 2D points are detected using a single image for marker-less pose estimation. There are methods for interest point detection using CNN. An example is SuperPoint~\cite{detone2018superpoint}, where 2D locations and descriptors are estimated for reliable performance of pipelines like Structure from Motion (SfM) or Simultaneous Localization and Mapping (SLAM), where stable and repeatable keypoints are required. By estimating the descriptors with SuperPoint the matching between 2D points is more easily tackled. Recently, SuperGlue~\cite{sarlin2020superglue} has been introduced for robust matching of interest points in a pair of images, using a Graph Neural Network. The interest points can be estimated with a CNN like SuperPoint, or other approaches.

Different planar marker methods for pose tracking provide highly accurate pose estimation compared to previously discussed approaches. However, the target's keypoints detection is typically carried out using classical computer vision techniques, which require lots of heuristics and parameter tuning for reliable performance under different environments. Recently, Deep ChArUco~\cite{hu2019deep} was proposed for pose estimation based on a planar marker that is robust in low lighting and high movement conditions. It estimates 2D corners in a sub-pixel accuracy with two CNNs. While being an attractive approach, it is focused on the pose estimation of a ChArUco board.

\section{Proposed Approach}
\begin{figure}[h]
    \centering
    \includegraphics[width=0.3\linewidth]{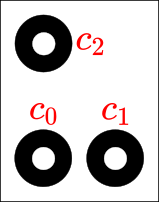}
    \caption{Fiducial marker used for pose estimation and the respective labels of each circle.}
    \label{fig:marker_labels}
\end{figure}

Here we describe MarkerPose, our proposed pose estimation approach. For that, we used a three-coplanar circle marker and a calibrated stereo vision system. The 3D coordinates of the centers define the 3D frame of the target. The circles, which due to perspective projection look like ellipses in the image, are labeled as shown in Fig.~\ref{fig:marker_labels}. These labels help us to solve the correspondences between both views for triangulation. But for that, we need to estimate these centers in a sub-pixel-level accuracy and classify them with the corresponding label or ID.

\begin{figure*}
    \centering
    \includegraphics[width=1\linewidth]{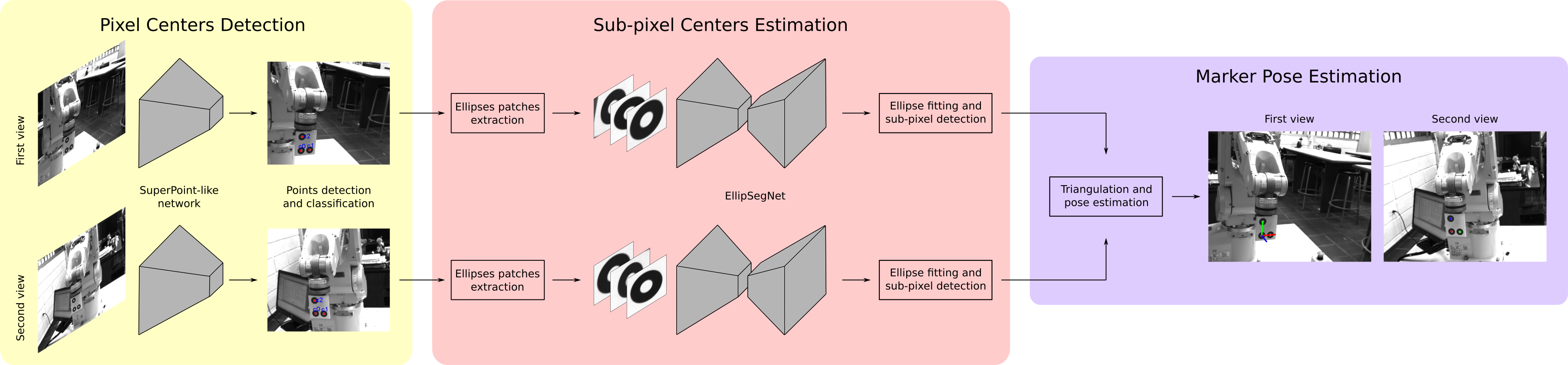}
    \caption{MarkerPose: the proposed approach for marker-based pose estimation, which consists of three stages. The first and second stage is for sub-pixel centers estimation, and the last stage for triangulation and target pose estimation.}
    \label{fig:method_pose}
\end{figure*}

The proposed MarkerPose method is shown in Fig.~\ref{fig:method_pose}, and consists of three stages. For pose estimation, we need first to detect the centers of the ellipses in a sub-pixel-level accuracy for both views. We perform center detection in the first two stages of our method: \textit{pixel centers detection} and \textit{sub-pixel centers estimation}. In the first stage, marker points in a pixel-level accuracy, and their IDs are estimated with a two-headed network based on SuperPoint~\cite{detone2018superpoint}. This SuperPoint variant is the same used in Deep ChArUco~\cite{hu2019deep}, where an ID-head replaces the descriptor head, and the same point localization head of SuperPoint is used. These classified rough 2D locations are used in the second stage to extract three square patches from each image around the pixel points that contain each ellipse. Then, we segment the ellipses to use their contour for sub-pixel centroid estimation. For that, we introduce EllipSegNet, a lightweight encoder-decoder network for fast ellipse contour estimation. Finally, in the last stage, with the sub-pixel matches of both views, triangulation is applied for 3D pose estimation. The correspondences are established with the points' IDs. Our method is robust to low lighting conditions and motion blur, essential for real-time applications and usage in different environments.

\subsection{Pixel Centers Detection}
\begin{figure*}
    \centering
    \includegraphics[width=1\linewidth]{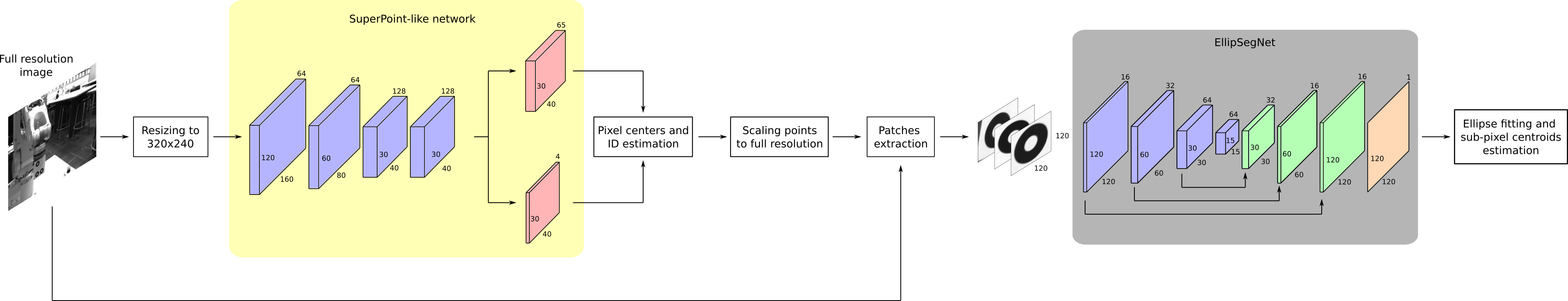}
    \caption{Sub-pixel centers detection for a single image.}
    \label{fig:method_detect}
\end{figure*}

Point detection from the first and second stages of our proposed method is shown in more detail in Fig.~\ref{fig:method_detect} for a single image. In this section, we focus on the first stage of MarkerPose which is based on a SuperPoint-like network. We use $1280 \times 1024$ images for training and testing. With the SuperPoint-like network, 2D pixel-level accuracy center detection is performed at a lower resolution of $320 \times 240$. For that, the input $1280 \times 1024$ images are resized to $320 \times 240$. The estimated rough 2D locations are then scaled-back to full resolution.

The original SuperPoint network~\cite{detone2018superpoint} consists of a backbone and two heads: a head for point localization and another for point descriptors. Instead of the descriptors head, an ID-head is used for point classification as shown Fig.~\ref{fig:method_detect}. The backbone is a shared encoder for both branches, which reduces the dimensionality of an image $W \times H$ to a factor of eight $W/8 \times H/8$, using eight convolution layers and three $2 \times 2$ max-pooling layers after every two convolutions. The point localization head produces a $W/8 \times H/8 \times 65$ tensor, which is a pixel location distribution of each $8 \times 8$ region of the input image and an additional channel for background or non-point locations. The ID head produces a $W/8 \times H/8 \times 4$ tensor, which is a classification distribution for each $8 \times 8$ region of the input image, where we have three labels for each center (see Fig.~\ref{fig:marker_labels}) and an additional label for background or non-center pixels. In this way, we have the following labels: 0 for $c_0$, 1 for $c_1$, 2 for $c_2$, and 3 for background. With the estimated IDs, we sort the resulting array of 2D points from 0 to 2 labels, where the first row corresponds to the $c_0$ point, and so on. With the output from this network, we solve the correspondence problem between both views.

The loss for this stage is the sum of the center location loss $\mathcal{L}_{loc}$, and the ID classification loss $\mathcal{L}_{id}$,
\begin{equation}
    \mathcal{L}(\hat{\mathbf{P}},\mathbf{P},\hat{\mathbf{C}},\mathbf{C}) = \lambda_1 \mathcal{L}_{loc}(\hat{\mathbf{P}},\mathbf{P}) + \lambda_2 \mathcal{L}_{id}(\hat{\mathbf{C}},\mathbf{C}) \enspace,
\end{equation}
where $\hat{\mathbf{P}}$ is the predicted $W/8 \times H/8 \times 65$ location tensor by the network, and $\mathbf{P}$ is the ground truth. $\hat{\mathbf{C}}$ is the predicted $W/8 \times H/8 \times 4$ classification tensor, and $\mathbf{C}$ the ground truth. For both $\mathcal{L}_{loc}$ and $\mathcal{L}_{id}$ cross-entropy loss is used.

\subsection{Sub-pixel Centers Estimation}
Once the circle center locations have been detected at a pixel-level resolution, in this second stage we refine these points to a sub-pixel-level accuracy for triangulation and accurate pose estimation. With the scaled-back pixel-level points from the SuperPoint-like network to the original resolution, we extract $120 \times 120$ patches around each detected center from the original $1280 \times 1024$ image. The ellipse is supposed within those patches for sub-pixel center estimation.

Since circles project as ellipses onto the image plane, we propose a lightweight encoder-decoder network called EllipSegNet for segmenting the extracted ellipse patches for contour estimation and sub-pixel centroid calculation as shown Fig.~\ref{fig:method_detect}. EllipSegNet is meant for real-time applications, and its weight is just 1.1 MB, and a forward pass takes 2.3~ms approximately with an NVIDIA Tesla P100. The encoder of the network consists of eight $3\times 3$ convolution layers with 16-16-32-32-64-64-64-64 channels, respectively. From the second convolution layer, three $2 \times 2$ max pool layers are used between each pair of convolutional layers. The decoder part of EllipSegNet consists of bilinear upsampling with a factor of 2 and skip connections via concatenation followed by two convolution operations. There are a total of six $3 \times 3$ convolution layers with 64-32-32-16-16-16 channels, respectively. Finally, a $1 \times 1$ convolution layer is used to map the 16 channels to 1 for binary segmentation.

The binary cross-entropy loss is used in this stage for the training of the network,
\begin{equation}
    \mathcal{L}(\mathbf{Y},\hat{\mathbf{Y}}) = - \sum_i \sum_j y_{ij} \log p(\hat{y}_{ij}) + (1-y_{ij}) \log [1- p(\hat{y}_{ij})]  \enspace,
\end{equation}
where $\hat{\mathbf{Y}}$ is the predicted mask, and $\mathbf{Y}$ the ground truth. Furthermore, $p$ is the sigmoid function,
\begin{equation}
    p(x) = \dfrac{1}{1+e^{-x}}  \enspace.
\end{equation}

For sub-pixel point estimation, we can compute the centroid of the white pixels of the mask with OpenCV's \texttt{moments} function, or we can use the contour of the ellipse to fit an ellipse with the OpenCV \texttt{fitEllipse} function. Either way, the detected center coordinates are estimated at sub-pixel accuracy. We adopt using \texttt{fitEllipse}, because we obtain more robust and accurate results than the other approach.

For sub-pixel estimation, the ellipse can be partially inside of the $120 \times 120$ patch. When there is more than one ellipse within the patch, the nearest to the image center is segmented. Although ellipse contour estimation is performed pixel-wise (discrete way), the sub-pixel calculation is not discrete, i.e., the estimation is not restricted to a discrete grid of positions, which enables higher accuracy. Furthermore, this sub-pixel estimation is carried out at full resolution, which leads to highly accurate pose estimation results because if we perform sub-pixel detection in a lower resolution, we need to scale-back the points to the original size. In this process, the small detection error in the lower resolution can be also scaled, and incur pose estimation errors. For example, a sub-pixel detection error less than a pixel at $320 \times 240$ can become larger than a pixel after scaling-back to the original resolution of $1280\times 1024$. Therefore, an error in low resolution can increase in the original resolution and this is critical in final pose estimation for high-accuracy applications.

\subsection{Marker Pose Estimation}
In the two previous sections, we described the sub-pixel 2D location estimation of the target's centers for a single image. This section describes the last stage of MarkerPose which consists of estimating the pose of the marker. This work is focused on pose estimation with a stereo vision system. With the estimated IDs, we solve the point correspondence problem. Hence, with the calibrated stereo vision system and the correspondences $\mathbf{x}_1 \leftrightarrow  \mathbf{x}_2$, we can recover the 3D position of the points. Following the camera pinhole model, the projection of a 3D point $\mathbf{X}$ in both image planes can be written as,
\begin{equation}
    \mathbf{x}_1 =  \mathbf{P}_1 \mathbf{X} \enspace, \qquad 
    \mathbf{x}_2 =  \mathbf{P}_2 \mathbf{X} \enspace,
\end{equation}
where $\mathbf{P}_1$ and $\mathbf{P}_2$ are the projection matrices for the cameras, estimated through calibration. To solve for $\mathbf{X}$, the homogeneous solution is given by
\begin{equation}
    \begin{bmatrix}
    x_1\mathbf{p}_1^{3\mathsf{T}} -  \mathbf{p}_1^{1\mathsf{T}}\\
    y_1\mathbf{p}_1^{3\mathsf{T}} -  \mathbf{p}_1^{2\mathsf{T}}\\
    x_2\mathbf{p}_2^{3\mathsf{T}} -  \mathbf{p}_2^{1\mathsf{T}}\\
    y_2\mathbf{p}_2^{3\mathsf{T}} -  \mathbf{p}_2^{2\mathsf{T}}\\
    \end{bmatrix} \mathbf{X} = \mathbf{0} \enspace,
\end{equation}
where $\mathbf{p}_1^i$ and $\mathbf{p}_2^i$ represent the $i$-th row of $\mathbf{P}_1$ and $\mathbf{P}_2$ respectively, as column vectors. We solve this with the OpenCV's \texttt{triangulate} function.

With the three 3D points of the target, we can define a 3D coordinate system. For that, we define the $c_0$ center as the origin of the target frame. The 3D position of this point gives us the translation vector $\mathbf{t}$ of the target frame with respect to the cameras. Furthermore, the circles $c_1$ and $c_2$ represent the direction of the $x$-axis and $y$-axis of the target coordinate system, respectively. The unit vector $\hat{\mathbf{x}}$ of the marker frame can be estimated with $c_0$ and $c_1$ 3D centers. Similarly, the unit vector $\hat{\mathbf{y}}$ with $c_0$ and $c_2$ 3D coordinates. Using cross product, the unit vector of the $z$-axis, that is perpendicular to the target, is calculated $\hat{\mathbf{z}}=\hat{\mathbf{x}} \times \hat{\mathbf{y}}$. Finally, the rotation matrix of the target frame relative to the stereo vision system is $\mathbf{R} = [\hat{\mathbf{x}} \;\: \hat{\mathbf{y}} \;\: \hat{\mathbf{z}}]$. In this way, we track the position $\mathbf{t}$ and orientation $\mathbf{R}$ of the target coordinate system. For accurate triangulation results, we compensate the 2D sub-pixel points for lens distortion before triangulation.

\subsection{Detection Method with a Classical Approach}
\label{sec:classical_method}
We also developed a center detection strategy using classical computer vision techniques. This method was designed to aid the labeling of training data and for comparison with the proposed approach using convolutional neural networks. This classical approach consists of binarizing and determining the image contours. As the outer black circle and internal white circles are concentric, their centers must have approximately the same image coordinates. We exploit this criterion to estimate the contour of the black circle and its center. This approach provides highly accurate results, but it is not robust. We need controlled environments for successful detection results. Furthermore, it requires parameter tuning to make it work in different environments and lighting conditions. Finally, as our stereo vision system is calibrated, we exploit epipolar constraints for solving the matching problem of points for pair of images, required for pose estimation.

\section{Implementation Details}
This section explains the details about the datasets, training of the SuperPoint-like and EllipSegNet networks, and the experimental setup.

\subsection{Training Dataset}
\begin{figure}[h]
    \centering
    \includegraphics[width=1\linewidth]{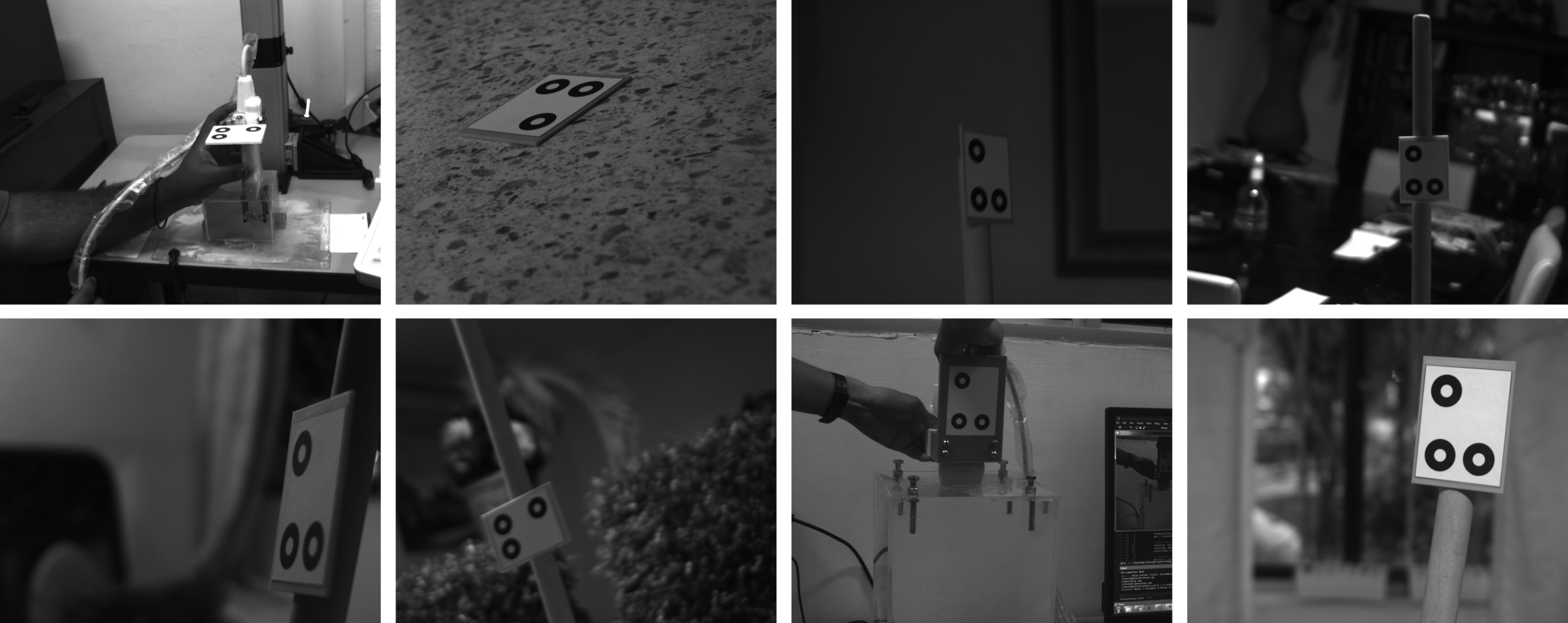}
    \caption{Example of training images for the SuperPoint-like network.}
    \label{fig:superpoint_training_example}
\end{figure}

We acquired a total of 5683 grayscale images with a $1280 \times 1024$ resolution. There are multiple positions and orientations of the target in our dataset, and the images were acquired at different lighting conditions. Furthermore, we also used two different target dimensions. Two example images are shown in Fig.~\ref{fig:superpoint_training_example}. For training the first stage of MarkerPose with the SuperPoint-like network, these images were resized to $320 \times 240$. For supervision, the center estimation was performed using the classical approach introduced in Section~\ref{sec:classical_method}. For data augmentation, we randomly applied the following operations: vertical and horizontal flip, random affine transformations, contrast modifications, motion blur, Gaussian blur, and Gaussian noise.

\begin{figure}[h]
    \centering
    \includegraphics[width=0.8\linewidth]{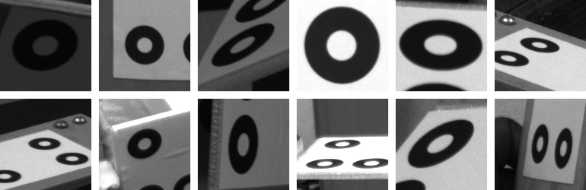}
    \caption{Example of training images for EllipSegNet.}
    \label{fig:patches_training_examples}
\end{figure}

For EllipSegNet training, we extract $120 \times 120$ patches from some of the images used for training the SuperPoin-like network, resulting in 11010 patches. Fig.~\ref{fig:patches_training_examples} shows some patch examples, where there are images where a single ellipse is in the patch and cases where we have more than one ellipse. We applied the same data augmentation transformations as described above for the keypoints detection network to ensure the robustness of the entire pipeline.

\subsection{Models Training}
We implemented the point detection and ellipse segmentation networks with PyTorch. A version was implemented with Python and another with C++ using LibTorch. For the SuperPoint-like network and the lightweight EllipSegNet, we used the Adam optimizer with a learning rate of 0.0001.

\subsection{Stereo Vision Setup}
The experimental stereo vision setup consisted of two monochromatic Basler cameras acA1300-200um (1280x1024, 203 fps), and two lenses Edmund Optics 58001 (12~mm fixed focal length). We calibrated our stereo vision system with OpenCV and a pattern of asymmetric circles. A total of 21 different poses of the pattern were acquired for calibration. We obtained a mean reprojection error of 0.1230~px and 0.1284~px for the first and second views, respectively. These errors are small and appropriate for accurate pose estimation.

\section{Experiments}
For validation, we compared MarkerPose with the classical approach discussed in Section~\ref{sec:classical_method}. For the experiments presented in this section, the distance between $c_0$ and $c_1$ centers of the target is 25~mm, and the distance between $c_0$ and $c_2$ centers is 40~mm.

\subsection{Validation Experiments with a Robotic Arm}
We evaluated the proposed pose estimation system with two experiments using an EPSON C3 robot (0.02~mm of precision). The robot was used for assessing the tracking accuracy. We attached the target to the arm's end effector, and we evaluated the displacement and rotation of the target moved by the robot.

\subsubsection{Displacement Evaluation}
\label{sec:dist_experiment}
\begin{figure}[h]
    \centering
    \includegraphics[width=1\linewidth]{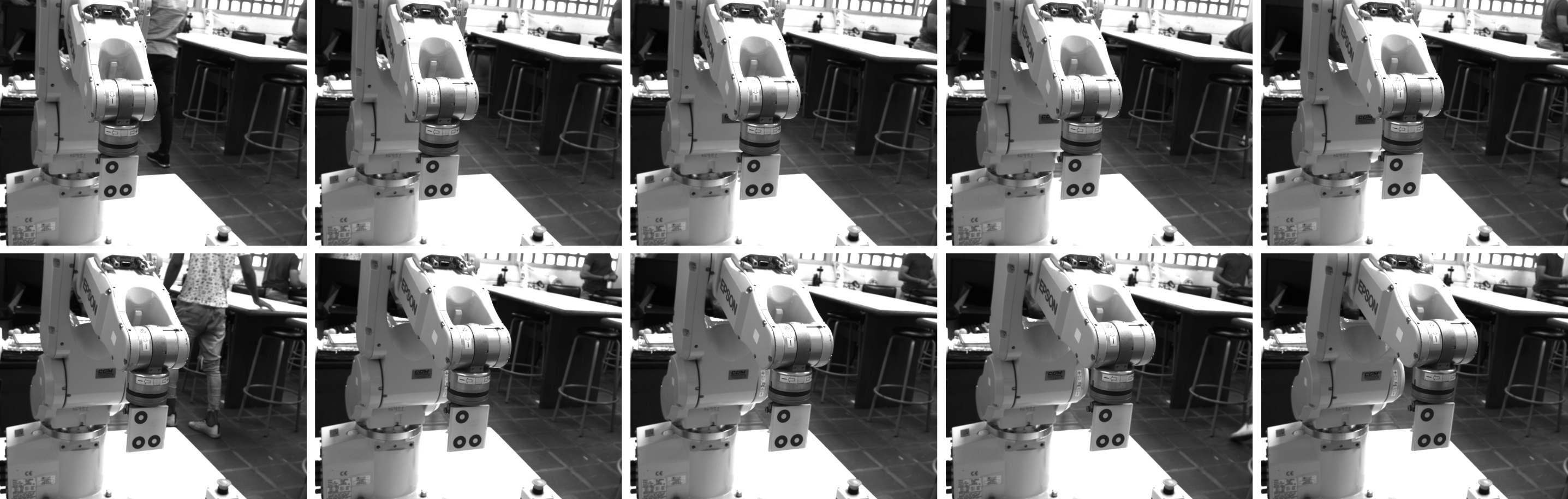}
    \caption{First 10 frames of the sequence for distance evaluation. A total of 20 positions were captured.}
    \label{fig:displacement_sequece}
\end{figure}

For this experiment, the end effector was displaced a fixed distance of $\Delta x = 10$~mm in a total of 20 positions. We measured the distance between consecutive frames, which give us 19 displacement estimations of each center of the target, i.e., a total of 57 displacement measurements. Fig.~\ref{fig:displacement_sequece} shows an example of the first 10 frames used for this experiment, acquired with the left camera (first view).

\begin{figure}[h]
    \centering
    \includegraphics[width=1\linewidth]{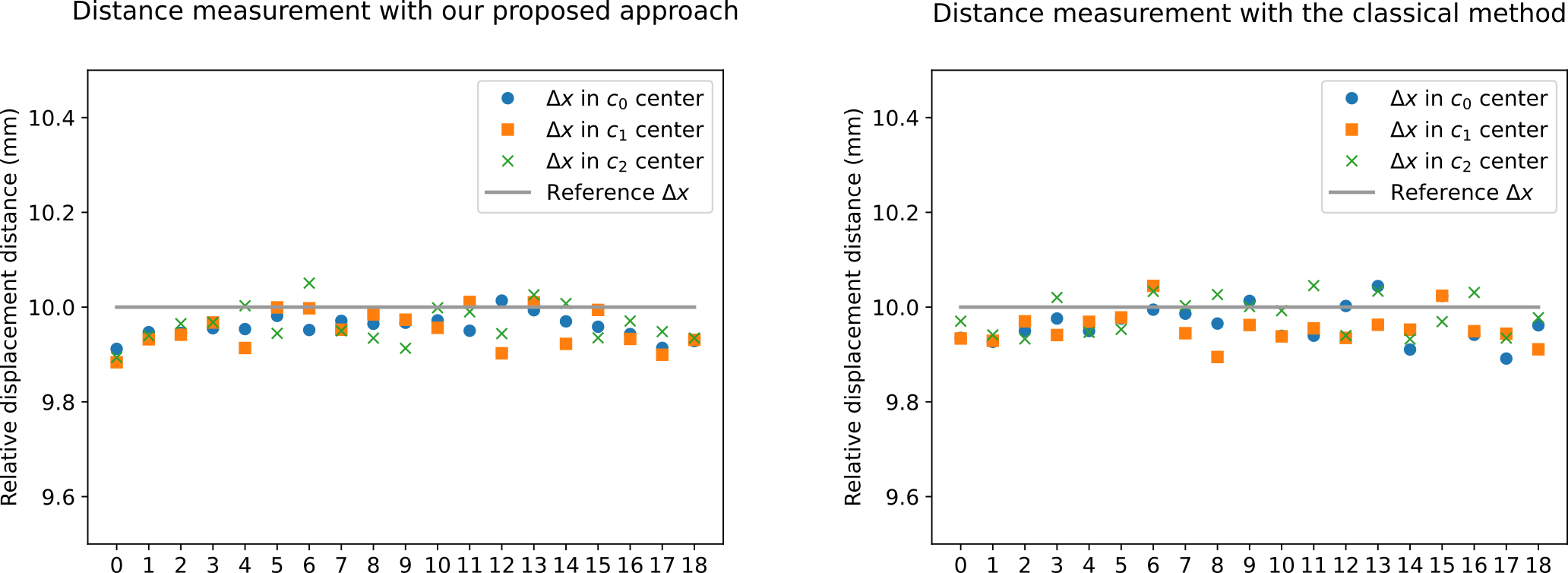}
    \caption{Displacement results from the 19 pairs of frames of MarkerPose and the classical method.}
    \label{fig:dist_results}
\end{figure}

Fig.~\ref{fig:dist_results} shows the results of the estimated relative distance between the 19 pairs of consecutive frames with MarkerPose and the classical approach. The measurements can be compared with the reference displacement line of 10~mm. Our proposed approach achieves high-accuracy results compared to the classical method.

We also estimated the mean absolute error (MAE), the RMS error, and the (maximum error) infinity norm $\ell_\infty$ of the errors of the calculated 57 distances of all centers, using both methods. The results are reported in Table~\ref{tab:dist_metrics}, where the results of MarkerPose with MAE and RMS metrics are comparable to the classical method. The maximum error is slightly larger with our approach, but a maximum error of 0.1164~mm is low enough for pose estimation in high precision applications.

\begin{table}[h]
\centering
\begin{tabular}{lccc}
\textbf{Method} & \textbf{MAE (mm)} & \textbf{RMS (mm)} & $\mathbf{\ell_\infty}$ \textbf{(mm)} \\ \hline
MarkerPose        & 0.0461            & 0.0544            & 0.1164                 \\
Classical       & 0.0446            & 0.0508            & 0.1086                 \\ \hline
\end{tabular}
\caption{Displacement experiment: performance metrics for the 57 distance measurements.}
\label{tab:dist_metrics}
\end{table}

Finally, we report the same metrics for the 3D distances between the centers estimated with the triangulated points. In this case, we have 20 measurements from the 20 acquired frames. These were obtained as the distance $d(c_0,c_1)$  between $c_0$ and $c_1$ centers and the distance $d(c_0,c_2)$  between $c_0$ and $c_2$ points, where $d(.)$ is the Euclidean distance between two 3D points. Table~\ref{tab:dist_metrics_points} shows the results. MarkerPose provides an overall better performance than the classical approach in this case.

\begin{table}[h]
\centering
\resizebox{\linewidth}{!}{%
\begin{tabular}{lcc|cc}
\textbf{}     & \multicolumn{2}{c|}{\textbf{MarkerPose}} & \multicolumn{2}{c}{\textbf{Classical method}} \\ \cline{2-5} 
              & $d(c_0,c_1)$             & $d(c_0,c_2)$            & $d(c_0,c_1)$            & $d(c_0,c_2)$            \\ \hline
\textbf{MAE (mm)}      & 0.1315                 & 0.2976               & 0.1558                & 0.3435                \\
\textbf{RMS (mm)}      & 0.1326                 & 0.2982               & 0.1569                & 0.3448                \\
$\mathbf{\ell_\infty}$ \textbf{(mm)} & 0.1692                 & 0.3257               & 0.1833                & 0.4107                \\ \hline
\end{tabular}%
}
\caption{Displacement experiment: metrics of the distance between the centers $c_0$ and $c_1$, and the centers $c_0$ and $c_2$, using the frames acquired for displacement estimation. Here, $d(.)$ is the Euclidean distance between two 3D points.}
\label{tab:dist_metrics_points}
\end{table}

\subsubsection{Rotation Evaluation}
\begin{figure}[h]
    \centering
    \includegraphics[width=1\linewidth]{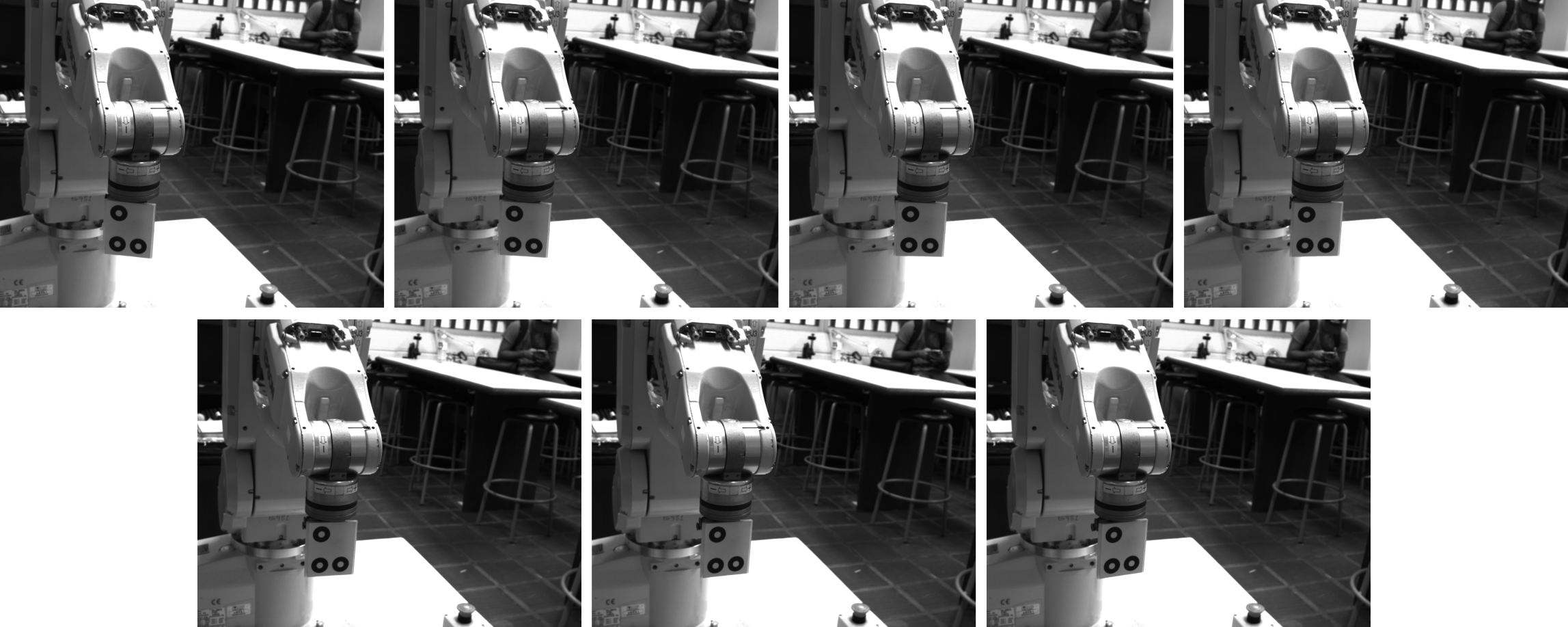}
    \caption{Entire sequence of 7 frames from the left camera for angle estimation.}
    \label{fig:rotation_sequece}
\end{figure}

The reference was established by rotating the robot's end effector at a fixed angle of $\Delta \theta = 5^\circ$ in a sequence of 7 frames. We measured the angle between pairs of consecutive frames, which gives us 6 rotation estimations. The calculation of the rotation angle between frames is performed with the following expression,
\begin{equation}
    \theta = \arccos \left[ \dfrac{\mathrm{Tr}(\mathbf{R}_{12})-1}{2}  \right]
    \enspace,
    \label{eq:ang_estimation}
\end{equation}
where $\mathbf{R}_{12}$ is the rotation matrix between the first and second frame, and can be estimated as $\mathbf{R}_{12} = \mathbf{R}_1^\mathsf{T} \mathbf{R}_2$. That is, $\mathbf{R}_{12}$ is the rotation matrix of the target coordinate system between frame 1 and frame 2. The entire sequence captured by the left camera is shown in Fig.~\ref{fig:rotation_sequece}.

\begin{figure}[h]
    \centering
    \includegraphics[width=1\linewidth]{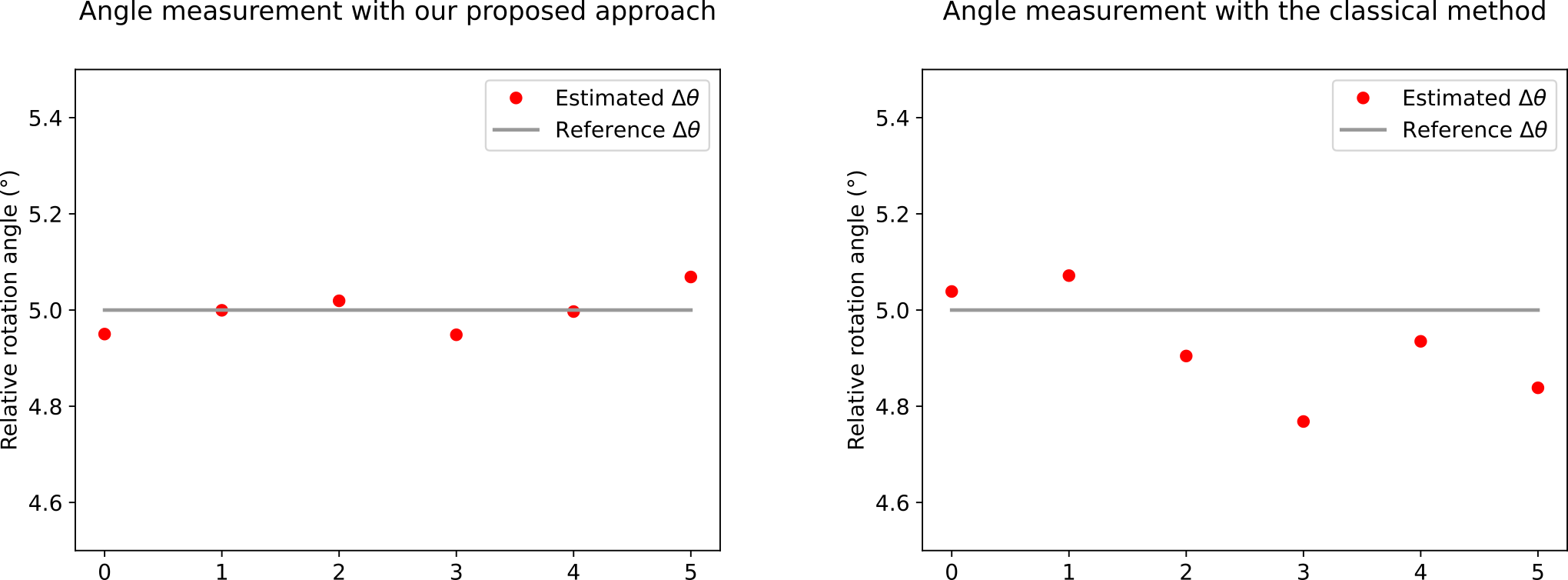}
    \caption{Rotation angle measurement results of MarkerPose and the classical method. A total of 6 pairs of consecutive frames were used.}
    \label{fig:rot_results}
\end{figure}

The estimated relative angle between pairs of frames is shown in Fig.~\ref{fig:rot_results} for MarkerPose and the classical method. By comparing with the reference line of 5$^\circ$, the proposed approach achieves better results than the classical approach with less dispersion. Furthermore, the metrics in Table~\ref{tab:rot_metrics} show the same behavior, where the MAE, RMS error, and the $\ell_\infty$ norm are significantly lower with MarkerPose. These results show the potential of our approach for highly accurate pose estimation applications.

\begin{table}[h]
\centering
\begin{tabular}{lccc}
\multicolumn{1}{c}{\textbf{Method}} & \textbf{MAE} & \textbf{RMS} & $\mathbf{\ell_\infty}$ \\ \hline
MarkerPose                            & 0.0322°      & 0.0413°      & 0.0687°                \\
Classical                           & 0.1107°      & 0.1290°      & 0.2317°                \\ \hline
\end{tabular}
\caption{Rotation experiment: angle metrics evaluation of the six pairs of frames.}
\label{tab:rot_metrics}
\end{table}

\begin{figure*}
    \centering
    \includegraphics[width=1\linewidth]{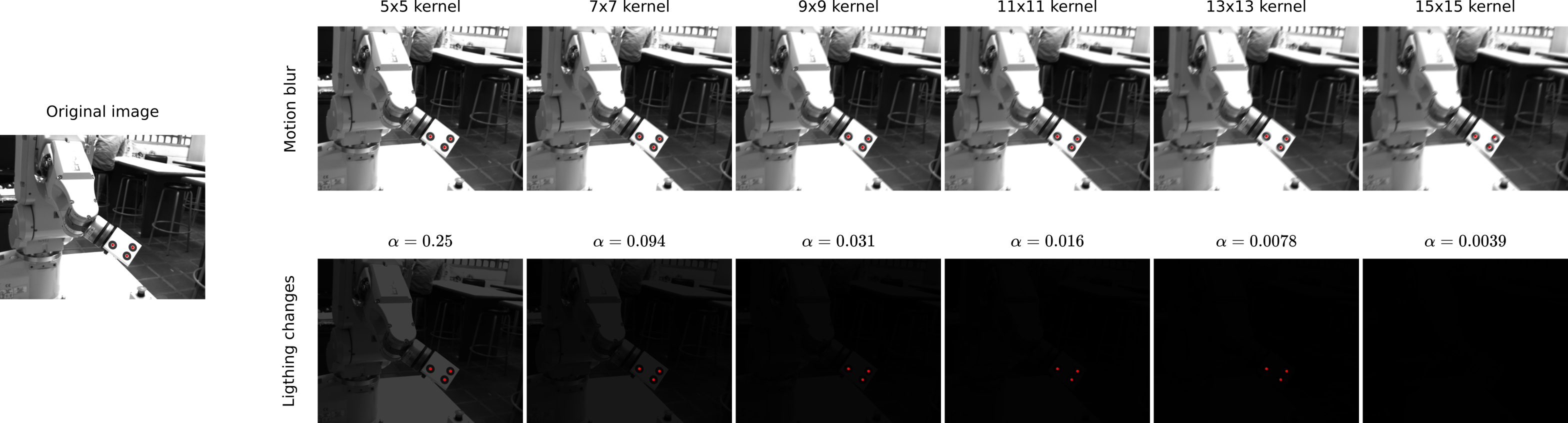}
    \caption{Results of our center point detection method for synthetically applied motion blur and lighting changes to a test input image.}
    \label{fig:synthetic_transform}
\end{figure*}

Similarly to the displacement experiment, the 3D Euclidean distance between centers is reported in Table~\ref{tab:rot_metrics_points}, where using the seven frames acquired, the distance between $c_0$ and $c_1$ centers and the distance between $c_0$ and $c_2$ centers, are estimated and the same metrics are used for these estimations. The metrics between the $c_0$ and $c_2$ centers are better with our proposed method. For the metrics between $c_0$ and $c_1$, the classical method is better except for maximum error.

\begin{table}[h]
\resizebox{\linewidth}{!}{%
\begin{tabular}{lcc|cc}
\textbf{}     & \multicolumn{2}{c|}{\textbf{MarkerPose}} & \multicolumn{2}{c}{\textbf{Classical method}} \\ \cline{2-5} 
              & $d(c_0,c_1)$             & $d(c_0,c_2)$            & $d(c_0,c_1)$            & $d(c_0,c_2)$            \\ \hline
\textbf{MAE (mm)}      & 0.1537                 & 0.2639               & 0.1315                & 0.3001                \\
\textbf{RMS (mm)}     & 0.1548                 & 0.2642               & 0.1416                & 0.3011                \\
$\mathbf{\ell_\infty}$ \textbf{(mm)} & 0.1896                 & 0.2867               & 0.2150                & 0.3387                \\ \hline
\end{tabular}%
}
\caption{Rotation experiment: metrics of the distance between the centers $c_0$ and $c_1$, and the centers $c_0$ and $c_2$, using the frames acquired for angle estimation.}
\label{tab:rot_metrics_points}
\end{table}

\subsection{Evaluation with Synthetic Effects}
We evaluated the proposed method under low lighting conditions and motion blur. The transformations were applied synthetically to the image, and the results are shown in Fig.~\ref{fig:synthetic_transform}. For motion blur, six different kernels sizes were applied to the original image, and center detection was carried out. Synthetic darkness was applied by scaling the original image with a factor $\alpha$. These $\alpha$ values range from 0.25 to 0.039. Those results show the robustness of our point detection method.

The distance estimation experiment was carried out again using the same dataset for different lighting changes with $\alpha$ values ranging from 0.5 to 0.0625. An $\alpha$ value of 1 does not represent a lighting change. The results are reported in Table~\ref{tab:lighting_evaluation}. These results show the robustness of MarkerPose to low lighting conditions, where even for the darkest case ($\alpha = 0.0625$) we obtain highly accurate results.

\begin{table}[h]
\centering
\begin{tabular}{lccc}
\textbf{$\alpha$} & \textbf{MAE (mm)} & \textbf{RMS (mm)} & \textbf{$\ell_\infty$} \\ \hline
1                & 0.0461            & 0.0544            & 0.1164                 \\
0.5              & 0.0451            & 0.0539            & 0.1352                 \\
0.25             & 0.0476            & 0.0557            & 0.1404                 \\
0.125            & 0.0528            & 0.0603            & 0.1313                 \\
0.0625           & 0.0568            & 0.0723            & 0.2130                 \\ \hline
\end{tabular}
\caption{Performance metrics for the distance estimation results for different lighting scaling values $\alpha$ from 0.5 to 0.0625. For $\alpha = 1$ there are no changes in the images.}
\label{tab:lighting_evaluation}
\end{table}

\subsection{Pose Estimation for a 3D Freehand Ultrasound System}
To showcase the highly accurate pose estimation result from MarkerPose, we describe an experiment with the 3D freehand ultrasound (US) technique~\cite{Huang:2017cd}. It consists of tracking the pose of a US probe while simultaneously acquiring US images. This procedure allows for mapping the 2D US scans to a volume by using a predefined transformation obtained via calibration~\cite{meza2020low}.

\begin{figure}[h]
    \centering
    \includegraphics[width=1\linewidth]{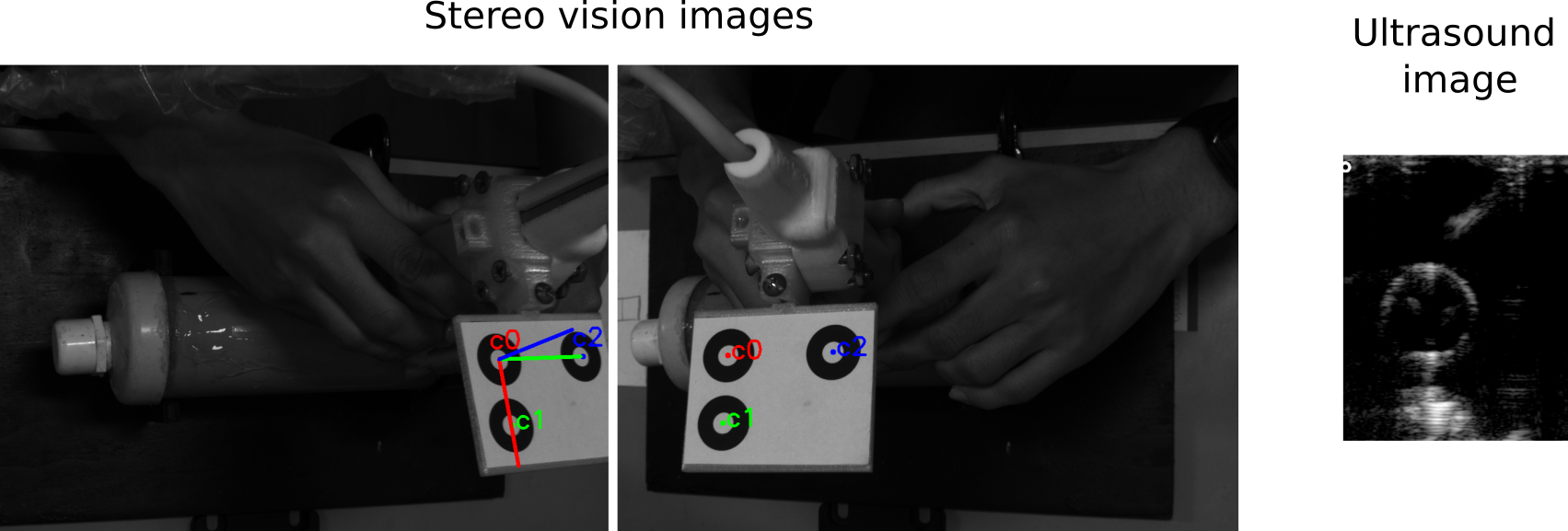}
    \caption{Example of 3D freehand ultrasound acquisition with the estimated pose with the MarkerPose method. The acquired ultrasound image shows a ring-like structure corresponding to the measured cylinder object.}
    \label{fig:3DfUS_acquisition}
\end{figure}

For this experiment, we used the same stereo vision setup and the US machine Biocare iS20. To evaluate our proposed pose estimation method, we reconstructed and estimated a cylinder's diameter with the US images. This cylinder was submerged in water for ultrasound visualization into another cylindrical object for easy acquisition. An example of stereo and US images during acquisition is shown in Fig.~\ref{fig:3DfUS_acquisition}, where the measured object is inside.

\begin{figure}[h]
    \centering
    \includegraphics[width=1\linewidth]{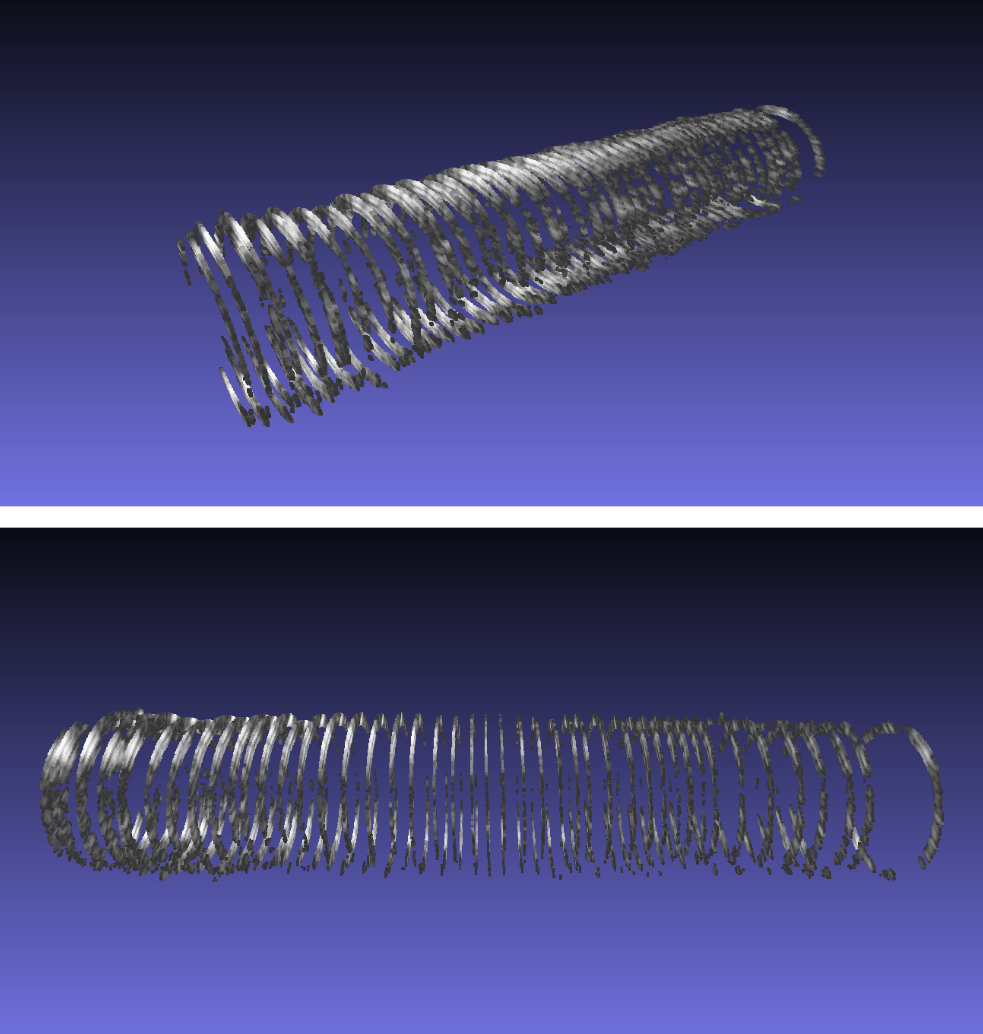}
    \caption{3D reconstruction results of the measured cylinder object with the 3D freehand ultrasound technique and the proposed pose estimation method. 94 2D ultrasound slices were used for 3D reconstruction.}
    \label{fig:3DfUS_cylinder_reconstruction}
\end{figure}

For this experiment, 94 ultrasound images were acquired along the central axis of the cylinder. We semi-automatically segmented the images for 3D reconstruction. The reconstruction results with our pose estimation method are shown in Fig.~\ref{fig:3DfUS_cylinder_reconstruction}, which is in agreement with the reference shape. Moreover, with this point cloud, we estimated the cylinder's outer diameter through least squares. For that, we use the outer points of the 3D coordinates. The estimated diameter with the point cloud was 15.999~mm, whereas the cylinder's diameter measured with a caliper was 15.82~mm. This result corresponds to an absolute error of 0.18~mm, which shows the method's suitability for highly accurate pose estimation applications.

\subsection{Runtime Evaluation}
We measured execution time for both of our point detection stages independently and for the entire MarkerPose pipeline. We use an NVIDIA Tesla P100 in a cloud computing platform in Python for runtime evaluation. Our first stage with the SuperPoint-like network takes 2.7~ms approximately for a single $320\times 240$ image. A forward pass in the second stage with the EllipSegNet runs in 2.3~ms approximately for a single $120\times 120$ array. Our entire pose estimation method, for two input $1280\times 1024$ images, including 2D points undistortion in both views, triangulation, and finally rotation matrix and translation vector estimation, runs at around 62 FPS. These results show the suitability of our proposed approach for real-time applications.

The same experiments were carried out with a GPU NIVIDA GTX 960M on a laptop using Python. A forward pass with the SuperPoint-like network runs at about 32.6~ms. A forward pass with the EllipSegNet takes 6.1~ms approximately. The entire pose estimation pipeline runs on average at 12 FPS. These results show that our method can achieve real-time performance even in an environment with fewer computational resources.

\section{Conclusions}
The need for highly accurate pose estimation is linked to applications that require reliable object tracking, like in the case of freehand ultrasound and other computer vision-based navigation and guidance systems. In most cases, the use of infrared or electromagnetic tracking devices leads to expensive systems which are not routinely used. Here, we have proposed MarkerPose: a real-time, highly accurate pose estimation method based on stereo triangulation using a planar marker detected with a deep neural network approach. Our multi-stage detection architecture leads to fast marker keypoint ID identification and sub-pixel keypoint detection at full image resolution by exploiting ellipse fitting using EllipSegNet, an introduced lightweight network. We showed the potential of the proposed method for accurate pose estimation and tracking under different lighting and motion blur conditions, and with a biomedical application. Future work involves designing novel markers that further exploit the capabilities of the proposed method.

\section*{Acknowledgement}
This work has been partly funded by Universidad Tecnologica de Bolivar project C2018P005 and Minciencias contract 935-2019. J.~Meza thanks Universidad Tecnologica de Bolivar for a post-graduate scholarship and MinCiencias, and MinSalud for a ``Joven Talento" scholarship.

{\small
\bibliographystyle{ieee_fullname}
\bibliography{egbib}
}

\end{document}